\documentclass[journal,twoside,web]{ieeecolor}
\usepackage{tmi}
\usepackage{cite}

\usepackage{amsmath,amssymb,amsfonts}
\usepackage{graphicx}
\usepackage{textcomp}

\usepackage{graphicx}
\usepackage{algorithm}
\usepackage[noend]{algpseudocode}
\usepackage[hidelinks]{hyperref}
\usepackage[capitalise]{cleveref}

\usepackage{caption}
\usepackage{subcaption}
\usepackage{booktabs}  
\usepackage{multirow}

\def\BibTeX{{\rm B\kern-.05em{\sc i\kern-.025em b}\kern-.08em
    T\kern-.1667em\lower.7ex\hbox{E}\kern-.125emX}}
\makeatletter
\renewcommand\thesection{\@arabic\c@section}
\renewcommand\thesubsection{\thesection.\@arabic\c@subsection}
\renewcommand{\@seccntformat}[1]{\csname the#1\endcsname.\ }
\def\thesectiondis{\thesection.}

\makeatother

\begin{document}

\title{
Minuscule Cell Detection in AS-OCT Images with Progressive Field-of-View Focusing
}

\author{
Boyu Chen\textsuperscript{\rm 1},
Ameenat L. Solebo\textsuperscript{\rm 2},
Daqian Shi\textsuperscript{\rm 1},
Jinge Wu\textsuperscript{\rm 1},
Paul Taylor\textsuperscript{\rm 1,*}\\
\small \textsuperscript{\rm 1}Institute of Health Informatics, University College London\\
\small \textsuperscript{\rm 2}Great Ormond Street Institute of Child Health, University College London\\
\thanks{This work was supported in module by an NIHR Clinician Scientist grant (CS-2018-18-ST2-005; ALS) and the AWS Doctoral Scholarship in Digital Innovation (awarded through the UCL Centre for Digital Innovation). }
\thanks{*Corresponding author: Paul Taylor.}
}

\maketitle
\begin{abstract}
Anterior Segment Optical Coherence Tomography (AS-OCT) is an emerging imaging technique with great potential for diagnosing anterior uveitis, a vision-threatening ocular inflammatory condition. A hallmark of this condition is the presence of inflammatory cells in the eye's anterior chamber, and detecting these cells using AS-OCT images has attracted research interest. While recent efforts aim to replace manual cell detection with automated computer vision approaches, detecting extremely small (minuscule) objects in high-resolution images, such as AS-OCT, poses substantial challenges: (1) each cell appears as a minuscule particle, representing less than 0.005\% of the image, making the detection difficult, and (2) OCT imaging introduces pixel-level noise that can be mistaken for cells, leading to false positive detections. To overcome these challenges, we propose a minuscule cell detection framework through a progressive field-of-view focusing strategy. This strategy systematically refines the detection scope from the whole image to a target region where cells are likely to be present, and further to minuscule regions potentially containing individual cells. Our framework consists of two modules. First, a Field-of-Focus module uses a vision foundation model to segment the target region. Subsequently, a Fine-grained Object Detection module introduces a specialized Minuscule Region Proposal followed by a Spatial Attention Network to distinguish individual cells from noise within the segmented region. Experimental results demonstrate that our framework outperforms state-of-the-art methods for cell detection, providing enhanced efficacy for clinical applications. Our code is publicly available at: \href{https://github.com/joeybyc/MCD}{https://github.com/joeybyc/MCD}.

\end{abstract}


\section{Introduction}

\begin{figure}[t]
\centerline{\includegraphics[width=\columnwidth]{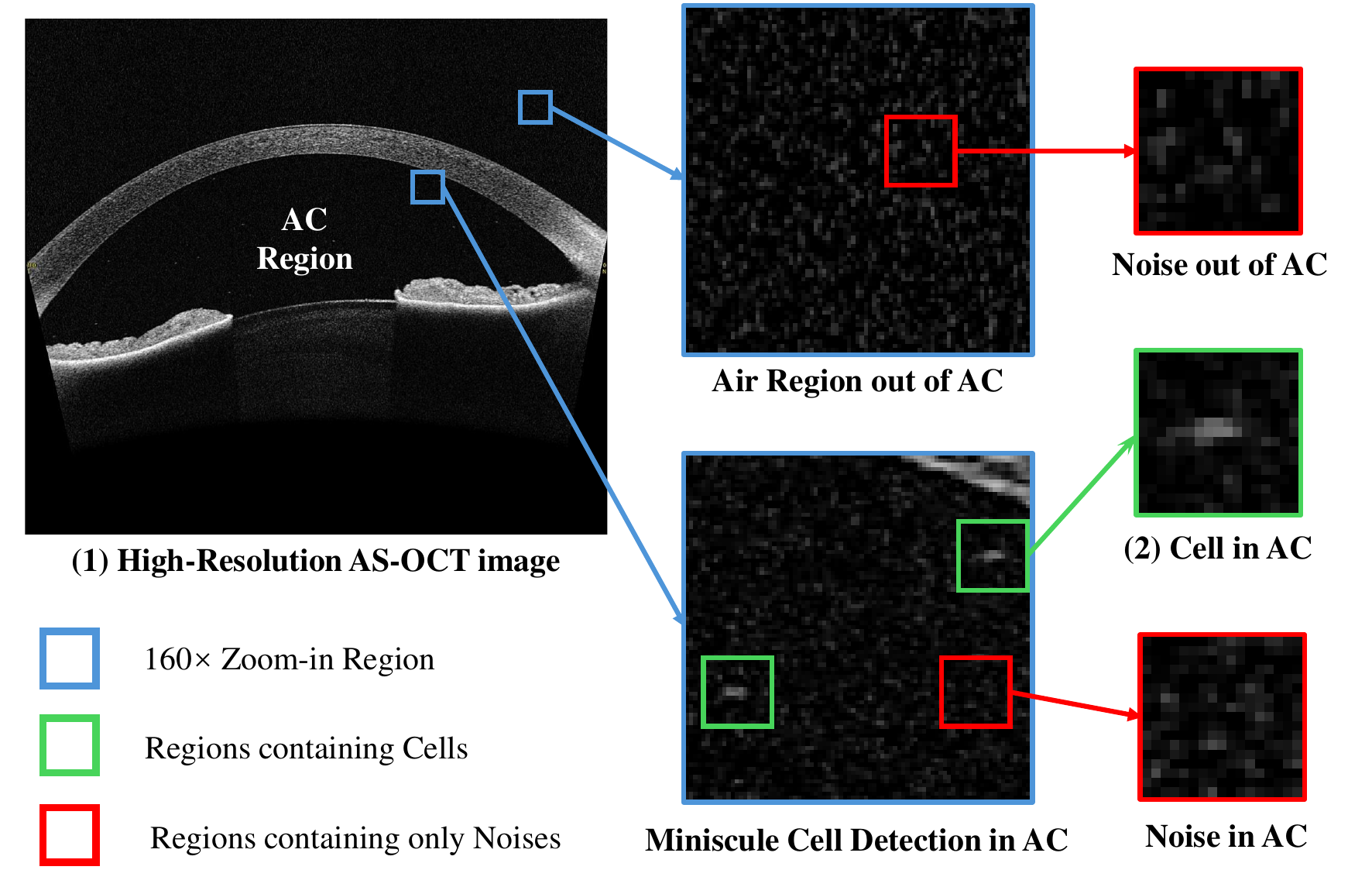}}
\caption{Minuscule inflammatory cells in the AS-OCT image.}
\label{fig:task}
\end{figure}

\label{sec:introduction}

Anterior Segment Optical Coherence Tomography (AS-OCT) has revolutionized the diagnosis and monitoring of ocular diseases, providing high-resolution, cross-sectional visualization of the anterior segment of the eye. There is particular scope for impact in anterior uveitis, a common inflammatory eye condition that can lead to significant vision impairment \cite{wakefield2005epidemiology, angeles2015characteristics, maleki2022pediatric, agrawal2012cataract, lee2014autoimmune, keane2014objective}. The hallmark of anterior uveitis is the presence of inflammatory cells in the anterior chamber (AC) of the eye \cite{standardization2005standardization}. Identifying these cells using AS-OCT images has attracted substantial research interest in recent years, as existing metrics are subjective and lack sensitivity\cite{igbre2014high, kumar2015analysis, invernizzi2017objective, akbarali2021imaging, tsui2022quantification, solebo2023anterior, sorkhabi2022assessment}.

However, this task is challenging due to the extremely small (minuscule) size of the cells. An AS-OCT image is shown in \cref{fig:task} (1), where we magnify two regions (inside and outside the AC) by 160 times and highlight them with blue boxes. Within these magnified regions, the hyperreflective particles enclosed in green boxes represent our target cells, as shown in \cref{fig:task} (2). These cells remain minuscule, with each cell occupying less than 0.005\% of the image - making them barely detectable to the naked eye. Moreover, OCT imaging introduces a substantial amount of pixel-level noise. This noise often closely resembles inflammatory cells in texture and color, making cell detection challenging. Unfortunately, traditional denoising methods cannot be applied as they would remove both noise and cells due to their similar appearance.

Traditionally, clinicians have relied on manual cell identification \cite{igbre2014high, kumar2015analysis, invernizzi2017objective, akbarali2021imaging, tsui2022quantification, solebo2023anterior}. As this is time-consuming, researchers have explored automated detection methods based on computer vision algorithms \cite{lu2020quantitative, etherton2023quantitative, agarwal2009high, rose2015aqueous, baghdasaryan2019analysis, keino2022automated, kang2021development, sharma2015automated, uthayananthan2025imaging}. These methods primarily employ thresholding techniques, in which pixels exceeding a specified brightness threshold are classified as cells. However, these approaches have significant limitations. First, determining an optimal threshold value is difficult: too high a threshold risks missing genuine cells, while too low a threshold may misidentify noise as cells. Second, these approaches still require human intervention to delineate regions to exclude false positives in irrelevant areas, introducing potential bias. Third, most studies bypass the validation of detections, instead directly using cell counts for subsequent analysis under the assumption that all detected objects are cells, negatively affecting their reliability.

Another potential solution for this task emerges from deep learning (DL)-based object detection methods, which have shown remarkable success in detecting objects in images \cite{ren2016faster, cai2018cascade, lin2017focal, yang2019reppoints, carion2020end, zhu2020deformable}. Among them, anchor-based approaches have gained widespread adoption. These methods work by hypothesizing numerous pre-set candidate regions (anchors) across the image and systematically examining each region for objects of interest. However, when applied to detect cells in high-resolution AS-OCT images, these approaches encounter significant limitations. On the one hand, the region proposals across the entire image produces an overwhelming number of candidates, dispersing focus away from the relevant areas. On the other hand, by down-scaling the original image or extracting global features before examining regions of interest, these methods lose their focus on crucial fine-grained details. This loss of attention to minute features makes minuscule objects virtually undetectable.

To that end, we propose a minuscule cell detection (MCD) framework based on a strategy of progressively focusing the field-of-view. Compared to direct detection across an entire high-resolution image, this strategy systematically refines the detection scope from the entire image to the target region where cells could be present (i.e., AC), and then to minuscule regions that potentially contain individual cells. Our MCD implements this strategy through two module:

The Field-of-Focus (FoF) module leverages a foundation vision model to zero-shot segment the AC region, focusing the search on the clinically relevant area. Within this region, the Fine-grained Object Detection (FOD) module uses a Minuscule Region Proposal to find minuscule regions potentially containing individual cells. Then, FOD uses a Spatial Attention Network to examine fine-grained details of these regions, distinguishing cells from noise. This progressive focusing approach not only preserves the fine-grained details crucial for cell detection but also reduces false positives by excluding irrelevant regions, overcoming the limitations of conventional threshold-based methods and existing DL-based methods.

The contributions of our paper are summarized as follows:

1) We propose a progressive field-of-view focusing strategy to detect minuscule inflammatory cells in AS-OCT images, where each cell represents less than 0.005\% of the image.

2) We develop the MCD to implement this strategy, featuring a FoF to zero-shot segment the AC region and a FOD to focus on fine-grained details to detect cells, and achieving state-of-the-art (SOTA) cell detection performance.

3) We reveal a critical limitation in previous threshold-based methods for cell detection in anterior uveitis studies. These methods may systematically underestimate cell populations, which provides new insights for future research.

4) Our MCD opens new possibilities for investigating the spatial distribution of inflammatory cells in anterior uveitis, potentially advancing our understanding of this disease.

\newpage

\begin{figure*}
\centerline{\includegraphics[width=\textwidth]{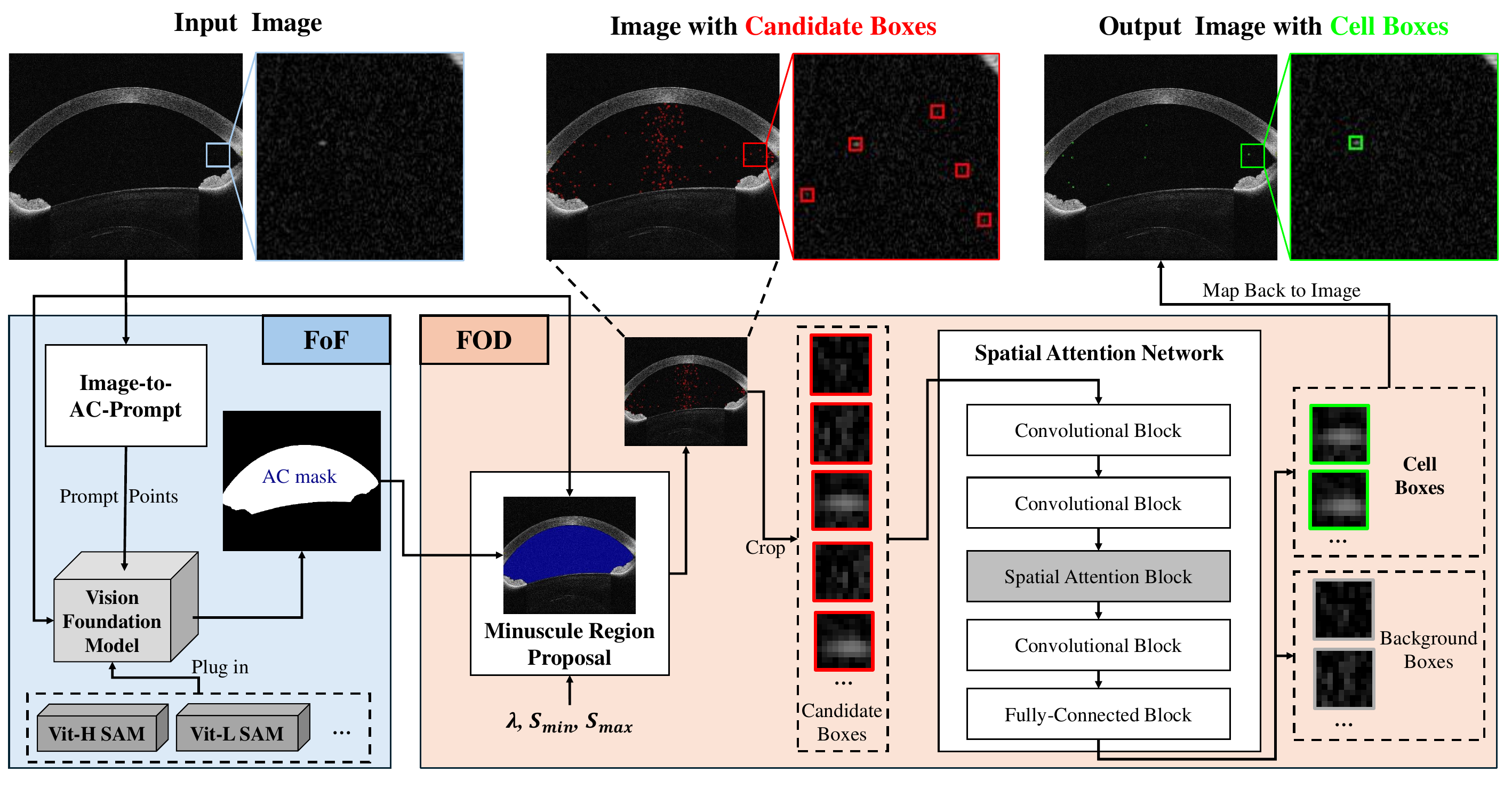}}
\caption{The overall architecture of our proposed minuscule cell detection framework.}
\label{fig:method}
\end{figure*}

\section{Related Works}
\label{sec:related_works}
\subsection{Finding Inflammatory Cells in AS-OCT Images}
The presence of inflammatory cells in the eye's anterior chamber (AC) is the primary diagnostic indicator for anterior uveitis. Clinicians traditionally employ slit-lamp biomicroscopic examination to detect and count these cells, translating their observations into the internationally standardized SUN grading system \cite{standardization2005standardization}. However, this approach is limited by inter-observer variability, subjective assessment, and poor sensitivity to minor changes in inflammation levels \cite{kempen2008interobserver, maring2022grading, wong2009effect}.

AS-OCT imaging has emerged as an alternative method for cell visualization, offering high-resolution imaging of the anterior segment. Early approaches involved manual cell identification in AS-OCT images \cite{igbre2014high, kumar2015analysis, invernizzi2017objective, akbarali2021imaging, tsui2022quantification, solebo2023anterior}, but this proved impractical for large-scale clinical applications due to its time-intensive nature.

To address these limitations, researchers began exploring automated detection methods. A natural starting point is threshold-based approaches, which are widely used in medical image analysis \cite{ridler1978picture, bradley2007adaptive, otsu1975threshold}. These methods classify pixels by comparing their intensity values (ranging from 0 for black to 255 for white) against a specified threshold - pixels exceeding the threshold are classified as bright objects, while those below it are considered background. Following this principle, researchers used these techniques for cell detection in AS-OCT images, with the Isodata algorithm \cite{ridler1978picture}, implemented as the default thresholding method in ImageJ\footnote{https://imagej.net/ij/}, becoming particularly prominent \cite{lu2020quantitative, uthayananthan2025imaging, etherton2023quantitative, agarwal2009high, baghdasaryan2019analysis}.

However, these methods face critical limitations in AS-OCT cell detection. First, threshold selection presents an inherent trade-off: high thresholds miss dimly visible cells, while low thresholds misidentify OCT imaging noise as cells. Second, these methods lack automatic detection capabilities for irregular-shaped AC regions, forcing researchers to either use rectangular image crops or rely on manual AC delineation. Third, they have not been validated for spatial accuracy of individual cell detections, instead directly using the detection results to test correlation with SUN scores or human counts. Our analysis reveals that threshold-based approaches can yield low recall rates, suggesting the underestimation of cell populations despite high correlations with clinical scores.

These limitations underscore the need for advanced cell detection methods in AS-OCT imaging—specifically, approaches that can accurately distinguish inflammatory cells from imaging artifacts while automatically identifying irregular-shaped AC regions and maintaining precise spatial detection accuracy.

\subsection{Image Segmentation}
As the cells only appear within the AC, segmenting this region is a crucial first step in our framework. DL segmentation models have had remarkable success in medical image segmentation \cite{zhu2024mp, ling2023mtanet, li2024scribformer, he2023h2former, zou2022graph}. However, their effectiveness depends on large-scale annotated training datasets, which is a significant limitation in medical imaging where expert annotations are costly and time-consuming to obtain. Vision foundation models present a compelling solution to this challenge \cite{kirillov2023segment, ma2024segment, gu2024lesam, jiang2024glanceseg, wu2023medical, zhang2023customized}. These models have been learned extensive prior knowledge from various segmentation tasks and can directly segment objects by taking user-provided prompts (clicked points or drawn boxes) along with the input image to automatically delineate the corresponding regions. However, existing vision foundation models for segmenting medical imaging still rely on either manual prompts during inference or task-specific fine-tuning. Unlike these approaches, our study achieves zero-shot AC segmentation without requiring training data, fine-tuning, or manual prompt inputs, making it more practical for real-world clinical applications.

\subsection{Object Detection}
DL-based object detection has achieved remarkable success in computer vision tasks, with anchor-based methods \cite{ren2016faster, cai2018cascade, lin2017focal} emerging as a popular approach. These methods place predefined anchor boxes across the image to serve as candidate regions. However, when applied to detecting minuscule cells in high-resolution AS-OCT images, the dense placement of anchors generates an overwhelming number of candidate regions, most of which are irrelevant, diluting the model's focus and increasing computational overhead. Alternative anchor-free approaches \cite{yang2019reppoints, carion2020end, zhu2020deformable} directly predict bounding boxes without preset anchors. However, both of these methods share a fundamental limitation when detecting minuscule objects: they rely on global feature extraction from downsampled images, inevitably losing the fine-grained details crucial for detecting objects that occupy less than 0.005\% of the image.

To address these challenges, we introduce a Minuscule Region Proposal strategy that specifically focuses on small-scale regions where individual cells might appear. We complement this with a Spatial Attention Network that maintains fine-grained feature extraction within these proposed regions. This combination preserves the subtle details necessary for accurate cell detection while efficiently managing computational resources by concentrating only on relevant regions.

\section{Method}
\label{sec:method}
Detecting inflammatory cells in high-resolution AS-OCT images is inherently challenging. To that end, we introduce a progressive field-of-view focusing strategy. The core idea is to gradually refine the focus of analysis: first, by concentrating on the AC region where such cells typically reside; then, by zooming in on extremely small, fine-grained subregions that may contain individual cells; and finally, by examining the spatial patterns embedded within the latent feature representations derived from these minuscule subregions to accurately distinguish cells from background noise. To operationalize this approach, we present a Minuscule Cell Detection (MCD) framework. As illustrated in \cref{fig:method}, our MCD comprises a Field-of-Focus (FoF) module that segments the AC region, followed by a Fine-grained Object Detection (FOD) module to identify individual cells within the segmented area.

The FoF (see the bottom-left of \cref{fig:method}) restricts the MCD's field-of-view from the entire image to only the AC region, where cells, if present, are confined. By automatically localizing the AC region, the FoF overcomes certain drawbacks associated with previous methods. In particular, threshold-based techniques struggle to eliminate false positives from irrelevant image areas without manual delineation, and existing DL-based detectors expend unnecessary computational resources analyzing the entire image, thus introducing additional false positives from regions outside the AC.

Conventional AC segmentation solutions typically rely on training DL models with extensive manual annotations. To circumvent this requirement, the FoF employs a zero-shot approach leveraging vision foundation models, which encode extensive prior knowledge of segmentation tasks. These models generate segmentation masks by processing both an image and given prompts (e.g., points within the target region, bounding boxes, or textual descriptions). To avoid manual prompt specification, we introduce an Image-to-AC-Prompt (I2ACP) algorithm, which automatically selects suitable prompt points from AS-OCT images. Notably, our FoF adopts a modular, plug-and-play design that permits seamless integration of different vision foundation models, thereby ensuring adaptability to future advances in this domain. Further details of the I2ACP are provided in \cref{sec:I2ACP}.

Within the AC region, the FOD further refines the focus by identifying minuscule regions—comprising less than 0.005\% of the entire image—that potentially contain individual cells. Existing DL-based detectors frequently examine large numbers of irrelevant proposals and lose critical fine-grained information during feature extraction, thus struggling to detect the cells. Observing that cells tend to appear as exceptionally small, locally brighter areas compared to their surroundings, we develop a Minuscule Region Proposal (MiRP) algorithm. The MiRP directly operates on high-resolution images to generate minuscule candidate boxes while preserving the subtle details necessary for cell detection. Further details of the MiRP are provided in \cref{sec:minuscule_proposal}.

Then, the FOD constrains the field of view even further by evaluating the spatial patterns encoded in the hidden feature maps extracted from each candidate box. While threshold-based methods struggle to distinguish between actual cells and background noise, we introduce a Spatial Attention Network (described in \cref{sec:spatial_attention}) that directly analyzes fine-grained spatial patterns within these minuscule regions. By integrating spatial attention mechanisms, this network adaptively highlights the most discriminative features essential for differentiating cells from noise. Ultimately, the network classifies each candidate box as either containing a cell or representing background noise (see the bottom-right of \cref{fig:method}), effectively eliminating false detections.

Through this progressive focusing strategy—transitioning from the entire image to the AC region, then to minuscule candidate areas, and finally to their inherent spatial characteristics—our MCD achieves robust and noise-resilient cell detection performance.

\begin{figure}[t]
\centerline{\includegraphics[width=\columnwidth]{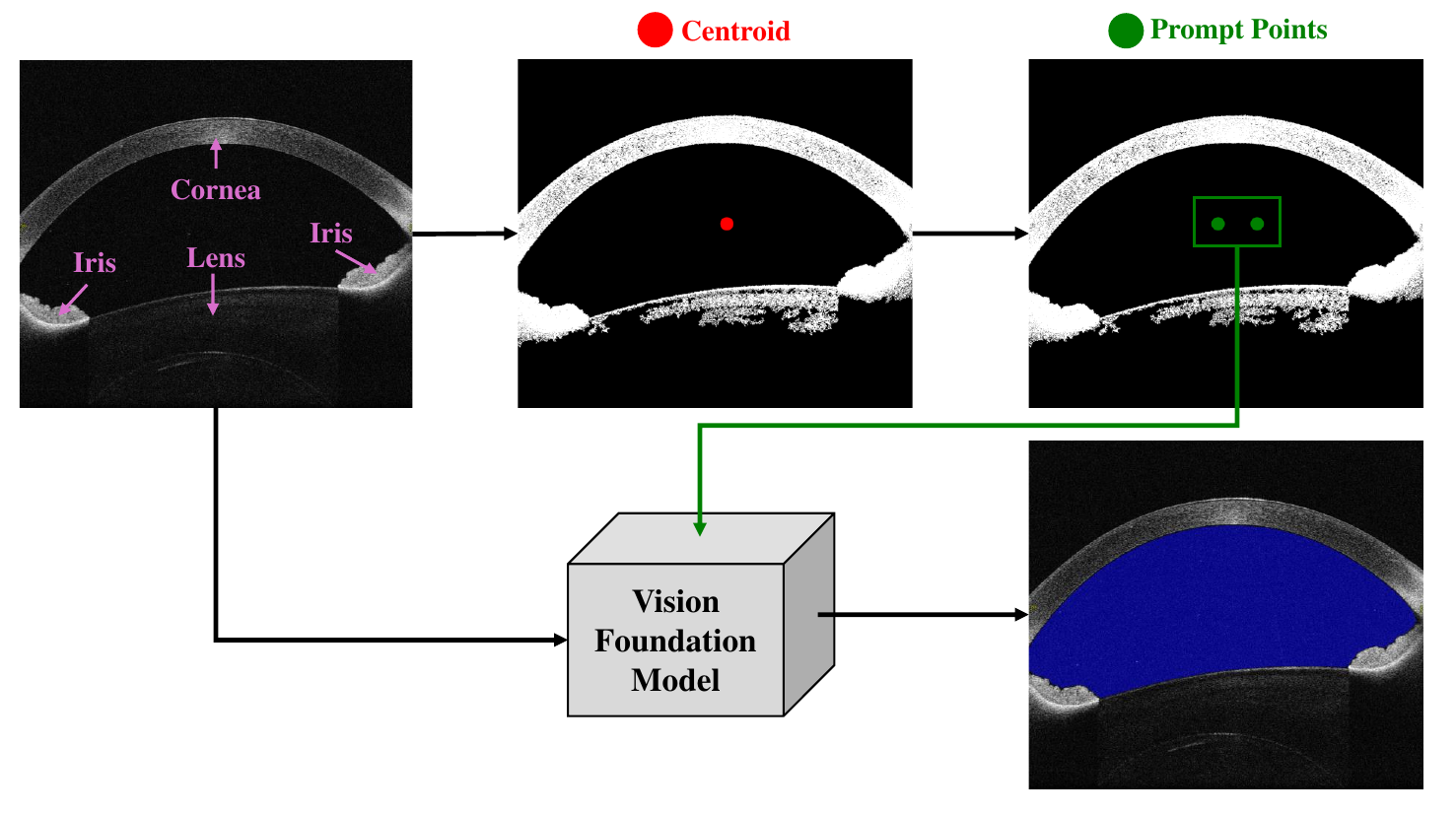}}
\caption{Using I2ACP to generate prompt points for supporting vision foundation model in segmeting AC region.}
\label{fig:I2ACP}
\end{figure}

\subsection{Image-to-AC-Prompt (I2ACP)}
\label{sec:I2ACP}
The Image-to-AC-Prompt (I2ACP) algorithm automatically generates prompt points to guide vision foundation models in AC region segmentation. This algorithm leverages a key anatomical observation: the anterior segment (comprising the cornea, iris, and lens) forms a natural boundary around the AC region, with its centroid consistently falling within the AC (see the top-left of \cref{fig:I2ACP}).

To locate this centroid, the I2ACP first converts the AS-OCT image to a binary mask by thresholding at the mean pixel intensity. In this mask, the largest connected component typically represents the anterior segment. In some cases, the anterior segment may appear split into
two components in the binary mask. 
Thus, the I2ACP examines the ratio of the areas between the second-largest and largest components. When this ratio exceeds a threshold $\mathcal{R}$, the I2ACP merges these components to form a complete anterior segment mask. The centroid ($x^{cent}$, $y^{cent}$) of this mask is then obtained (see the top-middle of \cref{fig:I2ACP}).
From this centroid, the I2ACP generates a set of prompt points $p_i$ by applying controlled offsets:
\begin{equation} \nonumber
p_{i} = (x^{cent} + \delta^{w}_{i}, y^{cent} + \delta^{h}_{i})
\end{equation}

These generated prompts (see the top-right of \cref{fig:I2ACP}), combined with the original image, are input to the vision foundation model to segment the AC region (see the bottom-right of \cref{fig:I2ACP}). The resulting AC mask ensures that subsequent cell detection in the FOD stage focuses exclusively on the clinically relevant region.

\subsection{Minuscule Region Proposal (MiRP)}
\label{sec:minuscule_proposal}

\begin{algorithm}
\caption{MiRP}\label{alg:region_proposal}
\begin{algorithmic}[1]
\Require Image $\mathcal{I}$, Focus mask $\mathcal{M}^{AC}$, Factor $\lambda$, min cell size $\mathcal{S}_{min}$, max cell size $\mathcal{S}_{min}$
\Ensure List of candidate boxes $\mathcal{B}$, each containing its top-left and bottom-right coordinates: $[(x_{tl}, y_{tl}), (x_{br}, y_{br})]$
\State $\mathcal{G} \gets \text{RGB2Gray}(\mathcal{I})$
\State $\mathcal{H} \gets \text{Histogram}(\mathcal{G}, \text{bins}=256)$
\State $\mathcal{P} \gets \mathcal{H} / \text{sum}(\mathcal{H})$
\State $\omega \gets \text{CumulativeSum}(\mathcal{P})$
\State $\mu \gets \text{CumulativeSum}(\mathcal{P} \times [0,1,\dots,255])$
\State $\mu_t \gets \mu[255]$
\State $\mathcal{T}^{init} \gets 0$
\State $\sigma^2_{max} \gets 0$
\For{$k \gets 0$ to $255$}
    \State $\omega_1 \gets \omega[k]$
    \State $\omega_2 \gets 1 - \omega_1$
    \If{$\omega_1 > 0$ and $\omega_2 > 0$}
        \State $\mu_1 \gets \mu[k] / \omega_1$
        \State $\mu_2 \gets (\mu_t - \mu[k]) / \omega_2$
        \State $\sigma^2 \gets \omega_1\omega_2(\mu_1 - \mu_2)^2$
        \If{$\sigma^2 > \sigma^2_{max}$}
            \State $\mathcal{T}^{init} \gets k$
            \State $\sigma^2_{max} \gets \sigma^2$
        \EndIf
    \EndIf
\EndFor
\State $\mathcal{T}^{opt} \gets \mathcal{T}^{init} \times \lambda$
\State $\mathcal{M}^{bright} \gets \mathcal{G} > \mathcal{T}^{opt}$
\State $\mathcal{C} \gets \text{ConnectedComponents}(\mathcal{M}^{bright})$
\State $\mathcal{C} \gets \mathcal{C} \land \mathcal{M}^{AC}$
\State $B \gets [ \, ]$
\For{each connected component $c_i$ in $\mathcal{C}$}
\If {$\mathcal{S}_{min} \leq \text{Area}(c_i) \leq \mathcal{S}_{max}$}
\State $(x_c, y_c) \gets \text{Centroid}(c_i)$
\State $x_{tl} \gets x_c - w/2$
\State $y_{tl} \gets y_c - h/2$
\State $x_{br} \gets x_c + w/2$
\State $y_{br} \gets y_c + h/2$
\State $\mathcal{B}.\text{append}([(x_{tl}, y_{tl}), (x_{br}, y_{br})])$
\EndIf
\EndFor
\State \Return $\mathcal{B}$
\end{algorithmic}
\end{algorithm}

We propose a Minuscule Region Proposal (MiRP) algorihtm to identify minuscule candidate boxes for potential cells while maintaining the fine-grained details of the high-resolution images. The key insight is that cells appear as bright, minuscule particles against a darker background in AS-OCT images. As detailed in \cref{alg:region_proposal}, the MiRP takes five inputs: a high-resolution image $\mathcal{I}$, a focus mask $\mathcal{M}^{AC}$ from the FoF, an adjustment factor $\lambda$, and valid cell size bounds $\mathcal{S}_{min}$/$\mathcal{S}_{max}$. The algorithm begins by converting the input image to a grayscale matrix $\mathcal{G}$ (line 1), where pixel values range from 0 (black) to 255 (white).

Then, the MiRP identifies a pixel-intensity $\mathcal{T}^{opt}$ to separate bright pixels from the darker background. First, the MiRP determines an initial pixel-intensity $\mathcal{T}^{init}$ (lines 2-18). This process, referring to Otsu's method \cite{otsu1975threshold}, begins by computing a histogram $\mathcal{H}$ of the grayscale image with 256 bins (line 2) and normalizes it to obtain a probability distribution $\mathcal{P}$ (line 3). The algorithm calculates two cumulative measures: $\omega$, representing the cumulative sum of probabilities (line 4), and $\mu$, representing the cumulative sum of the product between probabilities and intensity values (line 5). The total mean intensity $\mu_t$ is computed from $\mu[255]$ (line 6).

After setting $\mathcal{T}^{init}$ and maximum between-class variance $\sigma^2_{max}$ to zero (lines 7-8), the algorithm iterates through each intensity level $k$ from 0 to 255 (line 9). For each iteration, it computes the probabilities of the two classes (background and foreground) as $\omega_1$ and $\omega_2$ (lines 10-11). If both classes have non-zero probabilities (line 12), it calculates their respective means $\mu_1$ and $\mu_2$ (lines 13-14) and computes the between-class variance $\sigma^2$ (line 15). When this variance exceeds the current maximum $\sigma^2_{max}$, the algorithm updates both the maximum variance $\sigma^2_{max}$ and the $\mathcal{T}^{init}$ (lines 16-18).

To obtain the optimal $\mathcal{T}^{opt}$, MiRP applies an adjustment factor $\lambda$ (line 19) to modify the original $\mathcal{T}^{init}$. When $\lambda < 1.0$, it lowers $\mathcal{T}^{init}$, enabling the detection of cells with lower pixel intensities that might be missed by the original higher $\mathcal{T}^{init}$. This adjustment expands the set of candidate regions to include slightly dimmer areas that may contain cells, which is particularly crucial for AS-OCT images where cells often appear with varying intensities. The selection of optimal $\lambda$ will be described in \cref{sec:experiment}.

Using the $\mathcal{T}^{opt}$, MiRP generates a binary mask $M^{bright}$ (line 20) and identifies connected components representing potential cell regions (line 21). The algorithm applies the focus mask $\mathcal{M}^{AC}$ to retain only the components within the anterior chamber region (line 22). Each connected component then undergoes size validation using the bounds $[\mathcal{S}_{min}, \mathcal{S}_{max}]$ (line 25). 

For each valid component, the algorithm computes its centroid (line 26) and generates a bounding box centered at this point with minuscule width $w$ and height $h$ (lines 27-30). Finally, the algorithm returns a list of candidate boxes, each specified by its top-left and bottom-right coordinates: ($x_{tl}, y_{tl}$) and ($x_{br}, y_{br}$), respectively (lines 31-32).

\subsection{Spatial Attention Network}
\label{sec:spatial_attention}
\begin{figure}[!t]
\centerline{\includegraphics[width=\columnwidth]{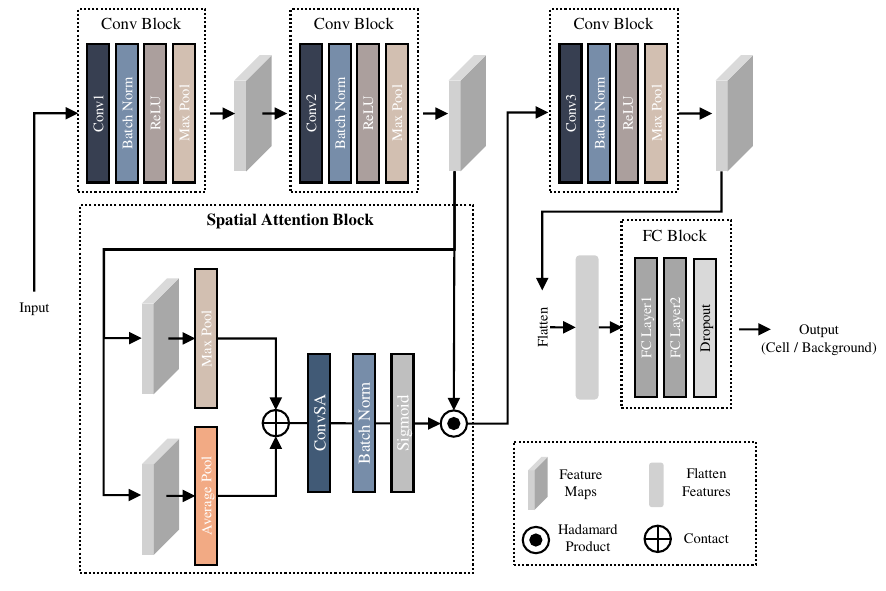}}
\caption{The architecture of our Spatial Attention Network.}
\label{fig:spatial_attention}
\end{figure}

The final focusing step examines spatial patterns within the minuscule regions to distinguish inflammatory cells from imaging noise. We achieve this through developing a Spatial Attention Network that adaptively focuses on discriminative spatial features while maintaining sensitivity to fine-grained details, as illustrated in \cref{fig:spatial_attention}.

The network begins with feature extraction through two consecutive convolutional blocks. Each block consists of a convolutional layer followed by batch normalization for stable training, ReLU activation for non-linearity, and max-pooling for spatial dimension reduction. This structure helps capture hierarchical features for cell identification.

The critical component for distinguishing cells from noise lies in the Spatial Attention Block, inspired by \cite{woo2018cbam}. Given an input feature map $\mathcal{F} \in \mathbb{R}^{C \times H \times W}$, this block generates an attention map $\mathcal{A} \in \mathbb{R}^{H \times W}$ by aggregating spatial information through dual pooling operations:
\begin{equation}\nonumber
\mathcal{A}(\mathcal{F}) = \sigma(\text{Conv}([\text{AvgPool}(\mathbf{F}); \text{MaxPool}(\mathcal{F})])))
\end{equation}
where $[\cdot;\cdot]$ denotes channel-wise concatenation and $\sigma(\cdot)$ is the sigmoid function. The average pooling captures global spatial context while max pooling identifies salient local features. The refined features are obtained through Hadamard product: $\mathcal{F} \odot \mathcal{A}(\mathcal{F})$.

Following the attention mechanism, the network employs another convolutional block with the same structure as previous blocks, followed by a fully connected block for final classification. The fully connected block flattens the feature maps and processes them through two dense layers with dropout to prevent overfitting, ultimately classifying each region as either cell or background.

During training, we construct training samples by cropping bounding boxes from annotated cell regions as positive samples (cells) and from non-cell areas as negative samples (background). To reflect the natural scarcity of cells in AS-OCT images, we maintain an imbalanced sampling ratio $\mathcal{N}^{pos} : \mathcal{N}^{neg}$ where $\mathcal{N}^{pos} < \mathcal{N}^{neg}$. The network is optimized using cross-entropy loss:
\begin{equation}\nonumber
\mathcal{L} = -\frac{1}{\mathcal{N}^{b}}\sum_{i=1}^{\mathcal{N}^{b}}\sum_{c=1}^{2} y_{i,c}\log(\hat{y}_{i,c})
\end{equation}
where $\mathcal{N}^{b}$ is the batch size, $y_{i,c}$ represents the ground truth label for class $c$, and $\hat{y}_{i,c}$ denotes the predicted probability. Training terminates if the validation loss shows no improvement for $\mathcal{N}^{patient}$ consecutive epochs.

\section{Experiment}
\label{sec:experiment}
\subsection{Datasets and Evaluation Metrics}
We evaluated our MCD using AS-OCT images from the Imaging in Childhood Uveitis studies \cite{akbarali2021imaging, etherton2023quantitative, solebo2024establishing} conducted at specialist children's and eye hospitals in London, UK. The study was conducted with approval from the UK's Health Research Authority (South Central Oxford B Research Ethics Committee, reference 19/SC/0283:SA02), and formal written consent was obtained from all participants.

We collected 1,376 high-resolution images with dimensions of $W \times H = 1598 \times 1465$ pixels. These images were divided into two non-overlapping sets for AC segmentation and cell detection evaluation. Six ophthalmic clinicians annotated the data via the Citizen Science project \cite{jones2018crowd} on the Zooniverse\footnote{https://www.zooniverse.org/}. All annotations underwent final verification by a senior ophthalmologist.

To evaluate AC segmentation performance, 630 images were annotated. We used the Intersection over Union (IoU) and the Dice coefficient as the metrics. IoU measures the overlap between the predicted and ground truth masks, while the Dice coefficient quantifies their similarity:
$$IoU = \frac{TP^{p}}{TP^{p} + FP^{p} + FN^{p}}$$
$$Dice = \frac{2 * TP^{p}}{2 * TP^{p} + FP^{p} + FN^{p}}$$
where $TP^{p}$ represents the overlap area between predicted and ground truth masks, $FP^{p}$ indicates the areas predicted as AC but not present in ground truth mask, and $FN^{p}$ represents the areas in ground truth mask that were missed in the prediction.

To evaluate cell detection performance, a different set of 746 images were annotated. Our clinicians annotated cells by clicking on them, with each click generating a 10 × 10 pixel bounding box centered on the clicked position, serving as the ground truth box. These minuscule boxes only occupy less than 0.005\% of the total image area. 

We evaluate both cell counting accuracy and spatial localization correctness. To determine a successful cell detection, we use two matching criteria. The first considers a prediction successful if the bounding box contains a ground truth cell point annotation, denoted by subscript ``point". The second criterion considers a detection successful if the IoU between predicted and ground truth boxes exceeds either 10\% or 30\% (denoted by subscripts ``10" and ``30", respectively). These relatively low IoU thresholds are chosen as small spatial displacements of compact bounding boxes can significantly decrease IoU values when detecting small objects \cite{cheng2023towards}. Once a ground truth cell is matched with a prediction, it is excluded from subsequent matching to prevent double-counting.

To evaluate cell counting accuracy, we calculate two Mean Absolute Error (MAE) metrics across all images: $MAE^{all}$ measures the average absolute difference between predicted and ground truth cell counts, and $MAE^c$ considers only correctly detected cells when computing the count difference.

To assess spatial localization accuracy, we use precision, recall, and F1-score. For all images, we define $TP^{b}$, $FP^{b}$, and $FN^{b}$ as the number of true positives (correct predicted boxes), false positives (predicted boxes without matched ground truth), and false negatives (unidentified ground truth boxes), respectively. These metrics are defined as:
$$Precision=\frac{TP^{b}}{TP^{b} + FP^{b}}$$
$$Recall=\frac{TP^{b}}{TP^{b} + FN^{b}}$$
$$F1=\frac{2*Precision *Recall}{Precision+Recall}$$

These metrics are computed separately for each matching criterion (`point', `IoU-10', and `IoU-30').

For each task dataset, the annotated images were randomly split into training, validation, and testing sets with ratios of 40\%, 10\%, and 50\%, respectively. The experiment was repeated 5 times, with different random splits of training, validation, and testing data in each iteration.



\subsection{Baselines and Implementation Details}

In the I2ACP, we empirically set $\mathcal{R}$ to 0.65 and generate two prompt points at offsets $(\delta^{w}_{1}, \delta^{h}_{1}) = (0, 0.1 \times W)$ and $(\delta^{w}_{2}, \delta^{h}_{2}) = (0, -0.1 \times W)$ from the anterior segment's centroid. In the MiRP, we set $\mathcal{S}_{max}=25$ based on clinical observations that larger objects are unlikely to be cells \cite{lu2020quantitative, etherton2023quantitative}, and $\mathcal{S}_{min}=1$ to include all potential cell candidates. Both $w$ and $h$ are set as 10. For our Spatial Attention Network, $\mathcal{N}^{pos}$, $\mathcal{N}^{neg}$, $\mathcal{N}^{patient}$, and $\mathcal{N}^{b}$ are set to 1, 5, 30 and 128, respectively. After training the Spatial Attention Network on the training/validation sets, we determine the optimal $\lambda$ through a parameter search by applying MCD with different $\lambda$ values to the validation set. The search ranges from 0.7 to 1.0 with a step of 0.01, and we select the $\lambda$ value that achieves the highest $F1_{point}$ on the validation set.

For the AC segmentation task, we compared our FoF with SOTA segmentation models: UNet \cite{ronneberger2015u}, UNet++ \cite{zhou2018unet++}, nnUNet \cite{isensee2021nnu}, Swin-Unet \cite{cao2022swin}, PSPNet \cite{zhao2017pyramid}, and DeepLabV3+ \cite{chen2018encoder}. We used Vit-B, Vit-L and Vit-H SAM \cite{kirillov2023segment} as our FoF's vision foundation model backbone to evaluate their performance.

For cell detection, we benchmarked against both threshold-based and existing DL-based approaches. To ensure a fair comparison, we only considered cells detected within the segmented AC region for these baselines to compare with the ground truth cell annotation.

For threshold-based baselines, we evaluated Otsu's method \cite{otsu1975threshold}, a widely-used approach for particle detection, and Isodata \cite{ridler1978picture}, which has been extensively applied for AS-OCT cell detection \cite{lu2020quantitative, etherton2023quantitative, agarwal2009high, baghdasaryan2019analysis, uthayananthan2025imaging}. Both methods generate binary masks by identifying connected components, and components larger than 25 pixels were also removed \cite{lu2020quantitative, etherton2023quantitative}. We created 10 × 10 pixel bounding boxes centered at each component's centroid as the predicted cell box. We further investigated the impact of filtering out components of different sizes, as smaller components could be noise \cite{lu2020quantitative, etherton2023quantitative}. We denote these variants as Otsu($\mathcal{S}_{min}$)/Isodata($\mathcal{S}_{min}$), where $\mathcal{S}_{min}$ represents the minimum pixel area for a connected component to be considered as a cell.

For DL-based baselines, we referenced a recent comprehensive evaluation of mainstream object detectors on small object detection \cite{cheng2023towards}. Based on their findings, we selected four SOTA models that outperformed others as our baselines: Faster-RCNN \cite{ren2016faster}, Cascade-RCNN \cite{cai2018cascade}, RetianNet\cite{lin2017focal}, and RepPoint\cite{yang2019reppoints}. Due to the extremely small size of cell bounding boxes (occupying less than 0.005\% of the image area), our initial attempts to train these models on full-resolution images failed to detect any cells. We then divided each image into 300 × 300 resolution patches for both training and inference. After detection, these boxes were mapped back to their corresponding locations in the original high-resolution images. For consistent evaluation, we standardized the detection results by converting each predicted box into a 10 × 10 pixel bounding box centered at the original box's center.

\begin{table}
\caption{Comparison of AC segmentation performance.}
\label{tab:res_chamber_seg}
\centering
\begin{tabular}{ccc}
\hline
\textbf{$Methods$} & \textbf{$IoU$(\%)}           & \textbf{$Dice$(\%)} \\
\hline
DeepLabV3+ \cite{chen2018encoder} & 93.49 & 96.23 \\
PSP Net \cite{zhao2017pyramid} & 93.34 & 96.41 \\
FoF (ViT-B SAM)& 95.28& 97.32\\
Swin-Unet \cite{cao2022swin} & 95.88 & 97.10\\
UNet++ \cite{zhou2018unet++} & 95.63 & 97.54 \\
 FoF (ViT-H SAM)& 95.72&97.28\\
UNet \cite{ronneberger2015u} & 95.98 & 97.86 \\
nnUNet \cite{isensee2021nnu} & 96.23 & 97.92 \\
FoF (ViT-H SAM) & \textbf{96.53} & \textbf{98.07} \\
\hline
\end{tabular}
\end{table}

\subsection{Results}
\cref{tab:res_chamber_seg} shows the IoU and Dice coefficient for the segmentation task. FoF with the Vit-H SAM backbone outperforms all baseline models, achieving a high IoU of 96.53\% and a Dice coefficient of 98.07\%.

\cref{tab:mae} compares the difference between predicted and ground truth cell counts. Our MCD achieves the lowest counting error across all metrics, with MAE$^{all}$, MAE$^{c}_{point}$, MAE$^{c}_{10}$, MAE$^{c}_{30}$ of 0.83, 0.71, 0.59, 0.82, respectively. This means that, on average, our MCD's cell count differs from the ground truth by less than one cell per image, indicating highly accurate quantification of inflammatory cells.

\cref{tab:prf} presents the precision, recall, and F1 scores under different matching criteria. Our MCD achieves the highest F1 scores across all criteria: 85.23\% for F1$_{point}$, 87.42\% for F1$_{10}$, and 82.61\% for F1$_{30}$. This represents significant improvements over both DL detectors and traditional threshold-based methods, showing our MCD's ability to effectively detect cells while maintaining a low false detection rate.

\begin{table}
\caption{Comparison Between Predicted and Ground Truth Cell Count}
\label{tab:mae}
    \centering
    \begin{tabular}{ccccc}
    \hline
         Methods&   MAE$^{all}$&MAE$^{c}_{point}$&  MAE$^{c}_{10}$&  MAE$^{c}_{30}$\\
    \hline
         Faster-RCNN \cite{ren2016faster}&   2.49&1.03&  0.90&  1.15
\\
         Cascade-RCNN \cite{cai2018cascade}&   2.45&0.86&  0.74&  0.97
\\
         RetianNet\cite{lin2017focal}&   1.74&1.12&  0.99&  1.24
\\
         RepPoint\cite{yang2019reppoints}&   2.88&1.18&  1.05&  1.29
\\
    \hline
         Otsu($\mathcal{S}_{min}=5$)&    2.70&2.77 &  2.72 &  2.85 
\\
         Otsu($\mathcal{S}_{min}=4$)&   2.33 &2.41 &  2.33 &  2.49 
\\
         Otsu($\mathcal{S}_{min}=3$)&   1.94 &2.02 &  1.94 &  2.12 
\\
         Otsu($\mathcal{S}_{min}=2$)&   1.59 &1.59 &  1.49 &  1.69 
\\
        Otsu($\mathcal{S}_{min}=1$)&  1.86 &1.14 & 1.03 & 1.25 
\\
    \hline
    Isodata($\mathcal{S}_{min}=5$)&   2.67&2.74 &  2.68 &  2.80 
\\
         Isodata($\mathcal{S}_{min}=4$)&   2.28 &2.36 &  2.29 &  2.44 
\\
         Isodata($\mathcal{S}_{min}=3$)&   1.90 &1.98 &  1.89 &  2.07 
\\
         Isodata($\mathcal{S}_{min}=2$)&   1.57 &1.56 &  1.46 &  1.66 
\\
        Isodata($\mathcal{S}_{min}=1$)&  1.99 &1.10 & 0.98 & 1.20 
\\
    \hline

        MCD&  \textbf{0.83} &\textbf{0.71} & \textbf{0.59} &\textbf{0.82}\\
    \hline
    \end{tabular}
\end{table}

\begin{table*}
\caption{Cell detection results on Precision (\%), Recall(\%) and F1(\%)}
\label{tab:prf}
    \centering
    \begin{tabular}{cccccccccc}
    \hline
         Methods&  Presicion$_{point}$&  Recall$_{point}$&  F1$_{point}$&  Presicion$_{10}$&  Recall$_{10}$&  F1$_{10}$&  Presicion$_{30}$&  Recall$_{30}$&  F1$_{30}$\\
    \hline
         Faster-RCNN \cite{ren2016faster}&  52.27&  78.56&  62.77&    54.06&   81.24&64.92&50.56&75.99&60.72
\\
         Cascade-RCNN \cite{cai2018cascade}&  54.93&  82.11&  65.82&    56.58&   84.58&67.80&53.33&79.71&63.9
\\
         RetianNet\cite{lin2017focal}
&  59.13&  76.66&  66.76&    61.28&   79.45&69.19&57.16&74.11&64.54
\\
         RepPoint\cite{yang2019reppoints}&  47.41&  75.47&  58.24&    49.13&   78.21&60.35&45.95&73.14&56.44
\\
    \hline
         Otsu($\mathcal{S}_{min}=5$)
&  \textbf{95.83}&  43.03 &  59.38 &    \textbf{98.46 }&   44.21 &61.01 &92.60 &41.58 &57.38 
\\
         Otsu($\mathcal{S}_{min}=4$)
&  95.22 &  50.59 &  66.06 &    98.06 &   52.10 &68.03 &92.17 &48.97 &63.95 
\\
         Otsu($\mathcal{S}_{min}=3$)
&  94.34 &  58.47 &  72.19 &    97.09 &   60.18 &74.30 &91.25 &56.56 &69.83 
\\
         Otsu($\mathcal{S}_{min}=2$)
&  91.22 &  67.35 &  77.48 &    93.96 &   69.38 &79.81 &88.37 &65.25 &75.06 
\\
        Otsu($\mathcal{S}_{min}=1$)& 74.29 & 76.70 & 75.46 &  76.44 &   78.92 &77.65 &72.09 &74.43 &73.23 
\\
    \hline
         Isodata($\mathcal{S}_{min}=5$)
&  95.65&  43.78 &  60.07 &    98.39 &   45.04 &61.79 &\textbf{92.67} &42.41 &58.19 
\\
         Isodata($\mathcal{S}_{min}=4$)
&  95.17 &  51.51 &  66.84 &    98.01 &   53.05 &68.84 &92.34 &49.98 &64.85 
\\
         Isodata($\mathcal{S}_{min}=3$)
&  94.08 &  59.38 &  72.81 &    96.88 &   61.16 &74.98 &91.10 &57.50 &70.50 
\\
         Isodata($\mathcal{S}_{min}=2$)
&  90.43 &  68.03 &  77.65 &    93.17 &   70.08 &79.99 &87.72 &65.99 &75.32 
\\
        Isodata($\mathcal{S}_{min}=1$)& 71.63 & 77.46 & 74.43 &  73.76 &   79.77 &76.65 &69.59 &75.26 &72.32 
\\
    \hline
        MCD& 85.40 & \textbf{85.16} & \textbf{85.23} &  87.64 &   \textbf{87.42} &\textbf{87.48} &82.77 &\textbf{82.54} &\textbf{82.61 }
\\
    \hline
    \end{tabular}
\end{table*}

\section{Discussion}
\label{sec:discussion}
\subsection{Zero-shot Segmentation}
Our FoF achieves superior AC segmentation performance through a fully automated approach that requires no annotated training data. This stands in contrast to the baseline models, which rely on manual annotations for training. This success lies in its integration of a vision foundation model with our I2ACP. This highlights the potential of prompt engineering to adapt vision foundation models to specialized domains. By leveraging domain-specific knowledge to generate prompts, we can unlock the capabilities of these models for anatomical structure segmentation without additional training overhead. This opens new avenues for other medical imaging tasks where annotated data may be scarce or costly to obtain.

\subsection{Detecting Minuscule Cells}
\begin{figure}[!t]
\centerline{\includegraphics[width=\columnwidth]{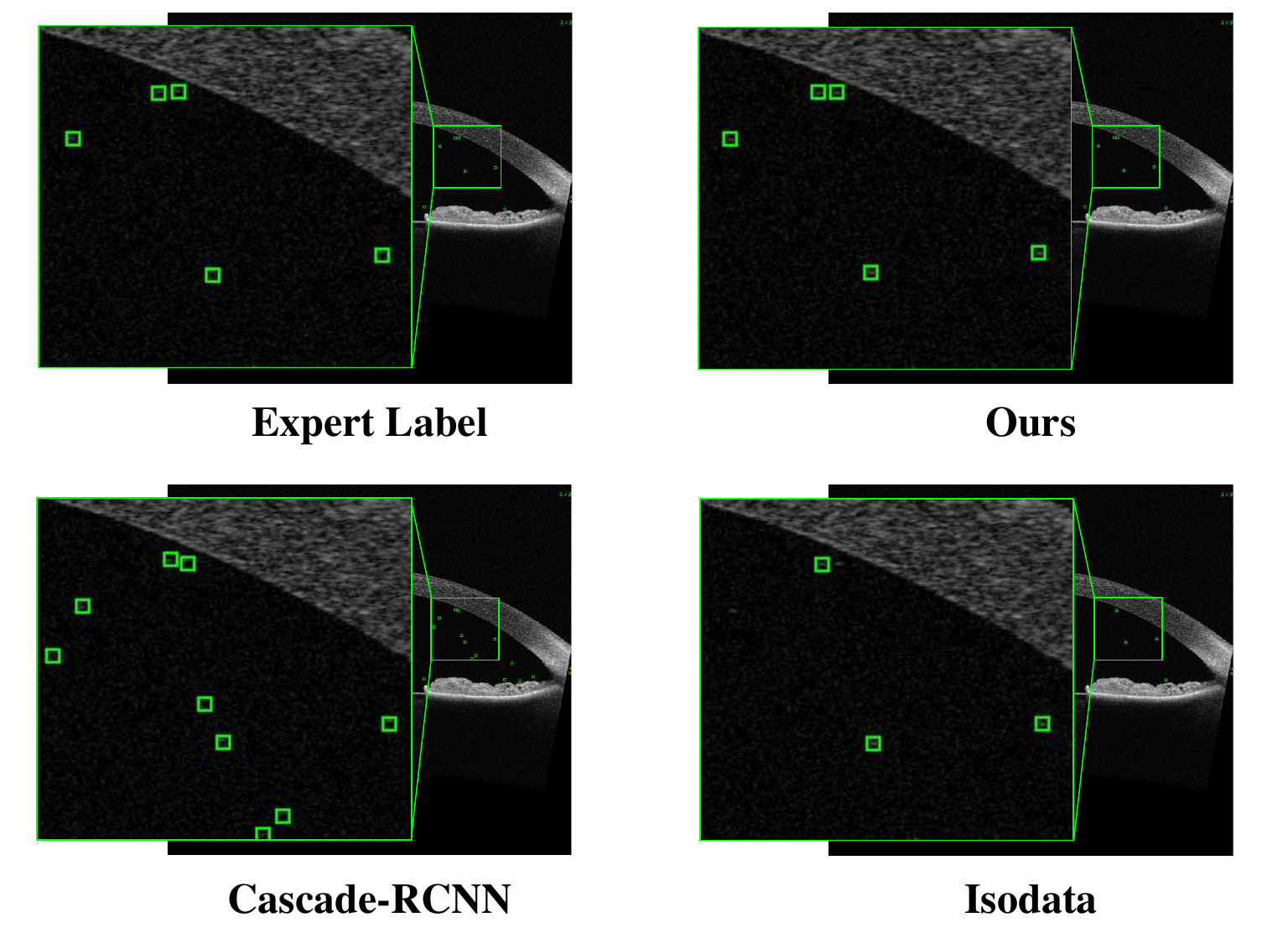}}
\caption{Qualitative comparison of cell detection results}
\label{fig:comparison}
\end{figure}

Our MCD demonstrates superior cell detection performance compared to existing approaches. While threshold-based methods with $\mathcal{S}_{min} < 4$ achieve better precision and F1 scores than DL detectors, MCD surpasses both approaches by delivering lower MAE scores in cell counting accuracy (\cref{tab:mae}); and higher Recall rates and F1 scores in spatial localization performance (\cref{tab:prf}). This superiority is further illustrated in \cref{fig:comparison}, where we qualitatively compare our MCD with expert annotations, a representative DL detector (Cascade-RCNN), and a conventional threshold-based method (Isodata with $\mathcal{S}_{min}=2$).

Initially, we trained the DL baselines on the original high-resolution images, but they failed to detect any cells. Even after dividing images into 300 × 300 pixel patches to mitigate the scale disparity between cells and the full image, these detectors still struggled with reliable detection. While DL detectors achieved higher recall than threshold-based methods, their lower precision resulted in inferior F1 scores, indicating frequent misclassification of noise as cells. As illustrated in \cref{fig:comparison}, Cascade-RCNN generates numerous false positive detections that deviate substantially from expert annotations. These limitations explain why previous studies predominantly relied on threshold-based approaches, which can directly process full-resolution images while maintaining better precision-recall balance. These findings highlight a critical limitation of current DL detectors in detecting extremely small objects within high-resolution medical images.

For threshold-based approaches, our analysis in \cref{tab:mae} and \cref{tab:prf} reveals an inherent trade-off between precision and recall. While these methods achieve high precision in cell detection, they suffer from low recall, potentially missing many cells, as visualized in \cref{fig:comparison}. With a strict filtering strategy ($\mathcal{S}_{min}=5$), where objects less than 5 pixels are considered noise, these methods achieve their highest precision but at the cost of severely compromised recall. Conversely, relaxing the threshold to $\mathcal{S}_{min}=1$ improves recall by considering smaller objects as potential cells, but substantially reduces precision. Notably, across all $\mathcal{S}_{min}$ settings, F1 scores remain consistently below 80\%, indicating these methods' fundamental limitation in accurately localizing cells within AS-OCT images.


In contrast, our MCD outperforms both threshold-based and DL-based approaches through its progressive field-of-view focusing strategy. As shown in \cref{tab:mae} and \cref{tab:prf}, MCD achieves high precision (85.40\% for Precision$_{point}$), high recall (85.16\% for Recall$_{point}$), and consequently a high F1-score (85.23\% for F1$_{point}$). Specifically, MCD surpasses the highest baseline F1-scores achieved by threshold-based methods, improving F1$_{point}$ by 7.58\% from 77.65\% of  Isodata($\mathcal{S}_{min}=2$), F1$_{10}$ by 7.49\% from 79.99\% of  Isodata($\mathcal{S}_{min}=2$), and F1$_{30}$ by 7.29\% from 75.32\% of  Isodata($\mathcal{S}_{min}=2$).

This balanced performance stems from the progressive focusing strategy. First, our MCD focuses the field-of-view on the clinically relevant AC region using the FoF module, effectively reducing the search space and minimizing false positives outside the AC. Then, our MCD refines its focus to extremely small regions through the Minuscule Region Proposal algorithm, concentrating computational resources where cells are most likely to appear. Finally, with the spatial attention mechanism, our MCD further refines its focus on spatial patterns embedded within the latent feature representations derived from these minuscule regions to accurately distinguish cells from background noise. This progressive refinement of focus from global to fine-grained detail enables MCD to achieve state-of-the-art performance in detecting minuscule inflammatory cells, as visually demonstrated in \cref{fig:comparison}, where our detections closely align with expert labels while avoiding both false positives and false negatives common to the baselines.

\subsection{Limitations}
Despite our MCD's superior performance in detecting minuscule cells in AS-OCT images, several limitations must be acknowledged. A challenge is the limited availability of public datasets for these tasks, making the development and release of comprehensive datasets essential for future research advances. Additionally, AS-OCT image quality can vary significantly - artifacts and quality degradation may occur \cite{patel2024quality}, potentially affecting MCD's effectiveness. In clinical settings, this challenge can be mitigated by implementing automated quality screening \cite{chen2023automated} to filter out low-quality images that provide minimal diagnostic value. Despite these limitations, we believe our MCD has great potential to be a fully automated tool for AC segmentation and cell detection in AS-OCT images.

\subsection{Clinical Implications}
Previous automatic detection methods used in anterior uveitis studies \cite{lu2020quantitative, etherton2023quantitative, agarwal2009high, rose2015aqueous, baghdasaryan2019analysis, keino2022automated, kang2021development, sharma2015automated} reported strong correlations between their detected cell counts and the SUN score (or manual count). However, these studies proceeded without verifying the correctness of individual cell detections. Our evaluation reveals two major limitations in these methods: cell count inaccuracy (as shown in \cref{tab:mae}) and imprecise spatial localization of detected cells (as shown in \cref{tab:prf}). While these methods demonstrated correlations with clinical scores, these correlations were likely based on incomplete and spatially inaccurate cell detection. This imprecision could negatively impact future research that requires accurate cell counting or studies investigating the relationship between spatial cell distribution patterns and disease progression.

In contrast, our MCD achieves good performance in detecting the cells. It opens up new possibilities for investigating the spatial distribution of inflammatory cells within the AC, which could enhance our understanding of disease pathogenesis and progression. Such insights could ultimately contribute to the development of more personalized treatment strategies, potentially improving patient outcomes. Furthermore, our progressive field-of-view focusing strategy could serve as a valuable reference for detecting extremely small objects in other medical imaging scenarios.

\section{Conclusion}
\label{sec:conclusion}
Detecting inflammatory cells in AS-OCT images, where each cell occupies less than 0.005\% of the image area, presents a critical challenge. Previous threshold-based methods have failed to accurately identify these minuscule cells, while DL object detection models struggle with such minuscule objects. To address this challenge, we introduce a progressive field-of-view focusing strategy implemented through our MCD framework. Our experimental results demonstrate that the MCD outperforms the SOTA baselines. Furthermore, this work reveals critical limitations in previous studies of anterior uveitis that relied on threshold-based methods for cell detection, suggesting potential underestimation of inflammatory cell populations. By providing more accurate cell detection and quantification, our framework enables enhanced monitoring of disease progression and treatment response in anterior uveitis. It creates new opportunities for investigating the spatial distribution patterns of inflammatory cells in AS-OCT images, potentially deepening our understanding of this disease. This advancement not only improves the technical aspects of cell detection but also holds promise for advancing clinical care and research in ocular inflammation.



\end{document}